\documentclass[conference]{IEEEtran}
\usepackage[utf8]{inputenc}
\usepackage[T1]{fontenc}
\usepackage{graphicx}

\usepackage[english]{babel}
\usepackage{caption}
\usepackage[absolute]{textpos}
\usepackage{newtxtext}   % replacement for times package
\usepackage{inconsolata} % changes the default typewriter font
\usepackage{xcolor}
\usepackage{hyperref}
\hypersetup{pdfborder={0 0 0},colorlinks=true,urlcolor=blue,linkcolor=blue,citecolor=blue,bookmarks=true}
\hypersetup{pdftitle={Approximate Nearest Neighbour Search on Dynamic Datasets: An Investigation}}
\hypersetup{pdfauthor={Ben Harwood, Amir Dezfouli, Iadine Chades, Conrad Sanderson}}

\usepackage{url}
\usepackage{balance}

\begin{document}

\title{Approximate Nearest Neighbour Search\\on Dynamic Datasets: An Investigation}

\author
  {
  Ben Harwood\textsuperscript{{\tiny~}$\dagger$},
  Amir Dezfouli\textsuperscript{{\tiny~}$\diamond$},
  Iadine Chades\textsuperscript{{\tiny~}$\dagger$},
  Conrad Sanderson\textsuperscript{{\tiny~}$\dagger\triangleleft$}\\
  ~\\
  \textsuperscript{$\dagger$}{\tiny~}\textit{CSIRO, Australia;}~
  \textsuperscript{$\diamond$}{\tiny~}\textit{BIMLOGIQ, Australia;}~
  \textsuperscript{$\triangleleft$}{\tiny~}\textit{Griffith University, Australia}
  }

\maketitle

\begin{abstract}
Approximate \mbox{k-Nearest} Neighbour (ANN) methods are often used for mining information
and aiding machine learning on large scale high-dimensional datasets.
ANN methods typically differ in the index structure used for accelerating searches,
resulting in various recall/runtime trade-off points.
For applications with static datasets,
runtime constraints and dataset properties 
can be used to empirically select an ANN method
with suitable operating characteristics.
However, for applications with dynamic datasets,
which are subject to frequent online changes (like addition of new samples),
there is currently no consensus as to which ANN methods are most suitable.
Traditional evaluation approaches do not consider the computational costs of updating the index structure,
as well as the rate and size of index updates.
To address this, we empirically evaluate 5 popular ANN methods 
on two main applications (online data collection and online feature learning)
while taking into account these considerations.
Two dynamic datasets are used,
derived from the SIFT1M dataset with 1~million samples and the DEEP1B dataset with 1~billion samples.
The results indicate that the often used \textit{k-d trees} method is not suitable on dynamic datasets
as it is slower than a straightforward baseline exhaustive search method.
For online data collection, the \textit{Hierarchical Navigable Small World Graphs} method
achieves a consistent speedup over baseline across a wide range of recall rates.
For online feature learning, the \textit{Scalable Nearest Neighbours} method
is faster than baseline for recall rates below 75\%.
\end{abstract}

\begin{textblock}{13.44}(1.28,14.75)
\hrule
\vspace{1ex}
\noindent
\scalebox{0.84}{\textbf{{$^\ast$}~Published in:} Lecture Notes in Artificial Intelligence (LNAI), vol.~15443, pp.~95--106, 2025.
~DOI:~\href{https://doi.org/10.1007/978-981-96-0351-0_8}{\tt 10.1007/978-981-96-0351-0\_8}}
\end{textblock}

\vspace{-1ex}
\section{Introduction}

Approximate k-Nearest Neighbour (ANN) search is a widely used technique
for computing local statistics in large datasets
comprised of high dimensional samples~\cite{Li_2020},
Application domains  include image retrieval, robotic localisation,
cross-modal search and various semantic searches~\cite{garg2021seqnet,prokhorenkova2020graph}. 

ANN methods are in contrast to brute-force search, 
where distances between each query and each sample in a dataset 
are exhaustively computed while maintaining a set of the k-nearest samples for each query.
While brute-force search provides exact results,
it has considerable computational costs, leading to long runtimes on large high-dimensional datasets.
To alleviate this, ANN methods can achieve sub-linear search times
by trading off between accuracy (recall) and runtime.

Within ANN search, an index structure is typically pre-computed (on a static dataset)
which is then used to accelerate the search process,
at the cost of some reduction in accuracy.
The type and configuration (tuning) of the index structure can be used to achieve a desired accuracy/runtime trade-off.
For example, graph based indexes can be configured to achieve high search accuracy~\cite{malkov2018hnsw},
while quantisation methods are better suited for faster searches with less exact results~\mbox{\cite{guo2020scann,Rama_1992}}.

To select an ANN method that is well-matched to application requirements,
the accuracy/runtime trade-off must be determined on a given dataset
for various configurations of each ANN method from a set of possible methods.
This can be computationally costly.
Several ANN evaluation frameworks have been established
to guide the selection and parameter tuning required
for achieving tractable searches~\cite{aumuller2017bench1,mlpack2023,matsui2020bench2}.

Recent applications, such as autonomous navigation,
require online data collection which results in an increasing number of samples over time, 
which in turn increases the computational cost of performing searches.
Furthermore, as the underlying dataset changes over time,
the index structure needs to be recomputed/updated
in order for ANN-based search to remain accurate and efficient.
Moreover, the need to update the index raises new questions,
such as when and how often it should be updated,
and how much data to use for each update.

\begin{figure}[!tb]
  \begin{minipage}{1\columnwidth}
    \begin{minipage}{0.05\textwidth}
      \centering
      \textbf{(a)} 
    \end{minipage}
    \hfill
    \begin{minipage}{0.85\textwidth}
      \centering
      \includegraphics[width=1\textwidth]{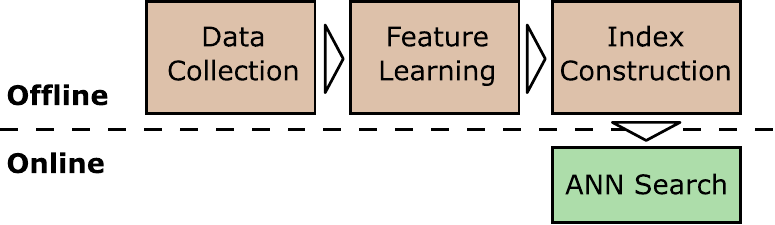}\\
    \end{minipage}
  \end{minipage}
  \vspace{1ex}
  \textcolor{gray}{\hrule}
  \vspace{1ex}
  \begin{minipage}{1\columnwidth}
    \begin{minipage}{0.05\textwidth}
      \centering
      \textbf{(b)} 
    \end{minipage}
    \hfill
    \begin{minipage}{0.85\textwidth}
      \centering
      \includegraphics[width=1\textwidth]{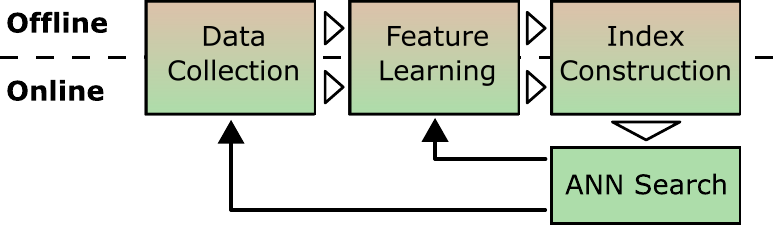}\\
    \end{minipage}
  \end{minipage} 
  \caption
    {
    \textbf{(a)}
    Traditional ANN evaluation approaches use a single batch of searches performed on a static index.
    \textbf{(b)}
    To better reflect ANN use on dynamic datasets,
    a more thorough evaluation must take into account index updates, as well as the rate and size of the updates.
    }
  \label{fig:pipeline}
  \vspace{1ex}
  \hrule
\end{figure}

A notable limitation of traditional ANN evaluation approaches
is the assumption that the index construction is a one-off offline process, 
as shown in Fig.~\ref{fig:pipeline}.
In turn this means that traditional approaches do not take into account
the computational costs of recomputing/updating the index structure for dynamic datasets,
and do not consider the rate and size of index updates.
Hence there is currently no consensus as to which ANN methods are most suitable for use on dynamic datasets.

To address this shortcoming,
in this work we evaluate%
\footnote
  {
  Associated source code is available at \url{https://github.com/data61/DyANN}.
  The code is extensible with new ANN methods
  and new categories of dynamic search problems.
  }
the appropriateness of 5~popular ANN methods
while explicitly taking into account the characteristics of search on dynamic datasets,
which includes the costs of index updates as well as the rate and magnitude of the updates.
The evaluations are performed on
two dynamic datasets that are representative of broad application areas.
The first dataset is modelled after real-time applications,
where each subsequent query is appended as a new sample to the dataset.
The second dataset represents online machine learning applications,
where samples are converging across incremental updates.

We continue the paper as follows.
Section~\ref{sec:ann_search_methods} briefly overviews the main families of ANN search methods.
Section~\ref{sec:characteristics} covers the characteristics of search on dynamic datasets.
Section~\ref{sec:experiments} empirically evaluates the covered ANN approaches on two dynamic datasets.
Concluding remarks and possible avenues for further research are given in Section~\ref{sec:conclusion}.

\vspace{1ex}
\section{ANN Search Families}
\label{sec:ann_search_methods}
\vspace{1ex}

In this work we consider 5 often used \mbox{ANN~methods} which have publicly available Python bindings:
Scalable Nearest Neighbours (ScaNN)~\cite{guo2020scann},
Inverted File index with Product Quantisation (IVFPQ)~\cite{johnson2019bf-ivfpq},
k-dimensional tree (k-d tree)~\cite{bentley1975kd,Rama_1992},
Approximate Nearest Neighbors Oh Yeah (ANNOY)~\cite{bernhardsson2013annoy},
\mbox{Hierarchical} Navigable Small World Graphs~(\mbox{HNSWG})~\cite{malkov2018hnsw}.
These methods can be placed into 3 broad families,
based on the type of index structure employed for speeding up searches:
(i)~quantisation, (ii)~tree, (iii)~graph.

{Quantisation} based ANN methods (such as ScaNN~\cite{guo2020scann} and IVFPQ~\cite{johnson2019bf-ivfpq})
cluster a dataset into local neighbourhoods where efficient residual distances are used.
Quantisation methods are related to hashing methods with their regular partition of a high dimensional space.
However, quantisation methods achieve this partition by grouping samples around cluster centers
rather than cutting up the space with hyperplanes.
Clustered samples are then mapped to a codebook using a reduced dimensional space within each cluster.
Cluster centers are learned from the intrinsic structure of a given dataset.

Tree based ANN methods (including k-d tree~\cite{bentley1975kd} and \mbox{ANNOY}~\cite{bernhardsson2013annoy})
connect dataset samples into a traversable index structure with a single source and no cycles.
Tree methods are frequently used for their fast index construction and searches.
Compared to quantisation methods,
tree methods have an additional memory cost of storing the edge information that enables the search paths.
Traversing a tree involves many local decisions
that use an increasingly smaller subset of the global information as sample dimensionality increases.
As such, higher search accuracy can only be achieved with a significant amount of backtracking on a single tree
or by searching in parallel over a forest of semi-redundant trees.
Despite these limitations,
standard k-d trees implementations continue to be applied to high dimensional search problems~\cite{kim2018useskd,xu2021useskd}.

Graph based ANN methods (such as HNSWG~\cite{malkov2018hnsw})
generalise tree methods by allowing any set of directed edges
that connect all dataset samples~\cite{harwood2016fanng,Shimomura_2021}.
Graph methods are some of the most recent and promising methods
due to favourable accuracy/runtime trade-offs in existing ANN benchmarks.
Similar to tree methods, they require the additional storage of edge information.
Graph methods are better suited for indexing large high dimensional data
due to their lack of an explicit hierarchical structure,
which ensures that all local regions are well connected to other areas of the graph.
However, the selection of an appropriate set of edges is computationally intensive,
leading to long index construction times.
The recent HNSWG approach~\cite{malkov2018hnsw}
avoids this cost by semi-randomly selecting a set of edges with desirable statistical properties.

The above ANN methods have parameters with recommended ranges 
that are typically obtained and tuned using static search problems~\cite{aumuller2017bench1,matsui2020bench2}.
As the nature of search on dynamic datasets is notably different (see Section~\ref{sec:characteristics}),
such recommended parameter ranges may not align
with the optimal range for dynamic search problems.

\vspace{1.5ex}
\section{Characteristics of Search on Dynamic Datasets}
\label{sec:characteristics}
\vspace{0.5ex}

We focus on two main application scenarios on dynamic datasets:
(i)~online data collection, and (ii)~online feature learning.
In online data collection, a dataset is progressively growing via addition of new samples.
Often the new samples are compared against existing samples in order to achieve detection of anomalies or estimation of saliency.
Within many machine learning (ML) processes,
the training of an embedding space~\cite{Girdhar_2023,Harandi_2016} is an example of online feature learning.
Updating ML model parameters during the learning process
will update the embedded representation of indexed samples.
This update can affect ANN index structures
and hence degrade the performance of subsequent searches.

In both scenarios, for ANN-based search to remain accurate and efficient on dynamic datasets,
the index structure needs to be recomputed after addition of samples or ML model updates.
We refer to sample addition as an \textit{addition event},
and an ML model update as an \textit{update event}.
The runtime cost of online index construction and updates is denoted as \emph{event processing time},
which is distinct from search runtime.

The index can be recomputed after the occurrence of each event, or after a number of events.
To take this into account, we define \emph{event batch size} as the number of events executed in a continuous block.
In a similar vein, we define \emph{search batch size} as the number of searches done in a continuous block.
In a straightforward evaluation, 
the processing of one batch type is followed by processing of the other batch type
(eg. block of events followed by block of searches).

\emph{Event rate} and \emph{search rate} define how many total events and searches, respectively,
are required in a fixed amount of time.
Given a fixed amount of computational resources,
we have a hard limit on the maximum number of operations
that can be shared across all events and searches in the available period of time.
As such, the required rate of events and searches is likely to have an impact on
the trade-off between search accuracy (recall) and search runtime.

\pagebreak
To avoid latency, the fixed budget of compute operations must be divided between each of the events
and searches processed during the available time period.
Increasing their rates will then result in a reduced amount of compute for processing individual events and searches.
For dynamic search problems, we expect that less compute per event
will eventually degrade the index quality.
Degraded index quality is likely to lead to slower search and/or lower search accuracy.
Furthermore, less compute per search is also likely to contribute to lower search accuracy.

The batching of events and searches can also impact the trade-off between accuracy and runtime in a less direct way.
For a fixed amount of compute in a given period of time,
let us consider a fixed number of events and searches that will be processed.
At each extreme of batching we can have events and searches alternating one after another
(eg.~an autonomous vehicle capturing and querying for each location as it travels),
or we can have a large chunk of events clumped together followed by a large chunk of the searches
(eg.~a closed loop learning pipeline completing a full forward pass
to then feeding back neighbourhood information about an entire dataset~\cite{Xin_2021}).

Larger batches of events can allow a total amount of compute
that is capable of applying global changes to an index structure,
while small batches might be limited to more localised changes.
We can expect that an index limited to localised changes is likely to degrade in quality,
thus reducing search accuracy and/or speed over time.

\vspace{1ex}
\section{Experiments}
\label{sec:experiments}

We perform a comparative evaluation
of the ANN methods overviewed in Section~\ref{sec:ann_search_methods}
on two dynamic datasets (described below)
representing online data collection and online feature learning.
All methods were limited to run on a single thread on AMD EPYC 7543 CPU,
using up to 10~GB RAM.
Restricting execution to a single thread aims to mitigate potential performance biases
that can occur at particular parallelisation thresholds.

As a baseline search method we use a straightforward technique
where distances between each query and each sample in a dataset are exhaustively computed
while maintaining a set of the k-nearest samples for each query;
a linear accuracy/runtime trade-off is obtained by computing distances to only a subset of samples;
there are no additional overheads for constructing or maintaining an index.

\subsection{Datasets}

The Online Data Collection (ODC) dataset is modelled after real-time applications
where each subsequent query is appended as a new sample to the dataset.
We use various subsets of the SIFT1M dataset~\cite{sift1m,Jegou_2011} which contains 1~million samples with 128 dimensions.
We use \emph{addition events} (defined in Section~\ref{sec:characteristics})
to grow each subset to double its initial size.
Initial samples are expected to be representative of the total sample pool,
and altering this may degrade the performance of some methods.

\pagebreak
The Online Feature Learning (OFL) dataset represents a generalised but flexible model for machine learning applications.
We apply \emph{update events} (defined in Section~\ref{sec:characteristics})
to model a dataset where samples are converging across incremental updates.
Various subsets of the DEEP1B dataset~\cite{babenko2016deep} are used,
which contains 1~billion samples with 96 dimensions.

\subsection{Comparative Evaluation}
\label{sec:comparative_eval}

For the ODC dataset, we start with 100K samples and perform 100K addition events,
with an event and search batch size of 1.
For the OFL dataset, we start with 5K samples and perform 100K update events,
with an event and search batch size of 200.
For each dataset, the evaluation was performed in two conditions:
\textbf{(a)}~parameters were tuned for each method
using the recommended ranges for static search problems,
and 
\textbf{(b)}~parameters were tuned for each method
via expanded parameter exploration in terms of the range and resolution of each parameter.
The recommended parameter ranges for static search problems were obtained from \cite{matsui2020bench2},
which in turn are based on the recommendations given for each 
evaluated method~\cite{bentley1975kd,bernhardsson2013annoy,guo2020scann,johnson2019bf-ivfpq,malkov2018hnsw}.

\begin{figure}[!tb]
  \begin{minipage}{1\columnwidth}
    \begin{minipage}{0.05\textwidth}
      \centering
      \textbf{(a)} 
    \end{minipage}
    \hfill
    \begin{minipage}{0.925\textwidth}
      \centering
      \includegraphics[width=1\textwidth]{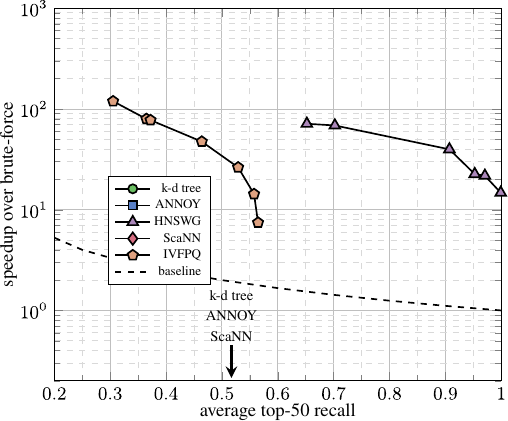}
    \end{minipage}
  \end{minipage}
  ~\\
  ~\\
  \begin{minipage}{1\columnwidth}
    \begin{minipage}{0.05\textwidth}
      \centering
      \textbf{(b)} 
    \end{minipage}
    \hfill
    \begin{minipage}{0.925\textwidth}
      \centering
      \includegraphics[width=1\textwidth]{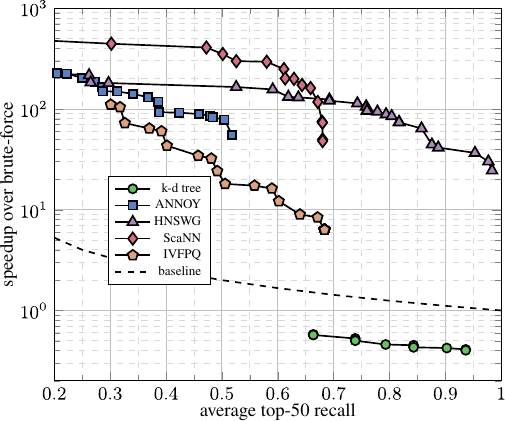}
    \end{minipage}
  \end{minipage} 
  \caption
    {
    Speedup (logarithmic scale) over brute-force search as a function of average top-50 recall, shown on logarithmic scale.
    The speedup is the ratio of time taken by brute-force search to the time taken by a given ANN method.
    Results obtained on the \textbf{Online Data Collection (ODC)} dataset,
    starting with 100K samples, followed by 100K addition events, with event and search batch size of~1.
    \textbf{(a)}~Parameters for each method were tuned following recommended practice for static search problems.
    \textbf{(b)}~Parameters for each method were tuned via expanded parameter exploration in terms of range and resolution.
    }
  \label{fig:odc_baseline_recall_runtime}
\end{figure}

\begin{figure}[!tb]
  \begin{minipage}{1\columnwidth}
    \begin{minipage}{0.05\textwidth}
      \centering
      \textbf{(a)} 
    \end{minipage}
    \hfill
    \begin{minipage}{0.925\textwidth}
      \centering
      \includegraphics[width=1\textwidth]{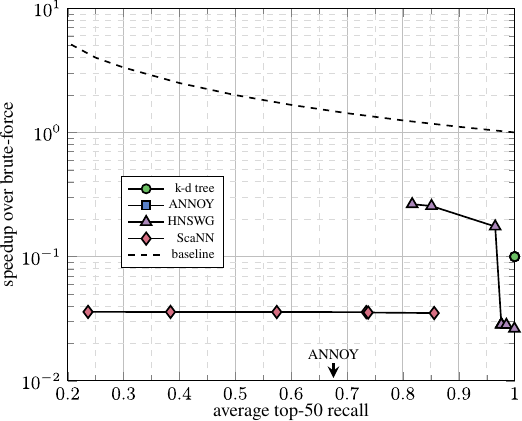}
    \end{minipage}
  \end{minipage}
  ~\\
  ~\\
  \begin{minipage}{1\columnwidth}
    \begin{minipage}{0.05\textwidth}
      \centering
      \textbf{(b)} 
    \end{minipage}
    \hfill
    \begin{minipage}{0.925\textwidth}
      \centering
      \includegraphics[width=1\textwidth]{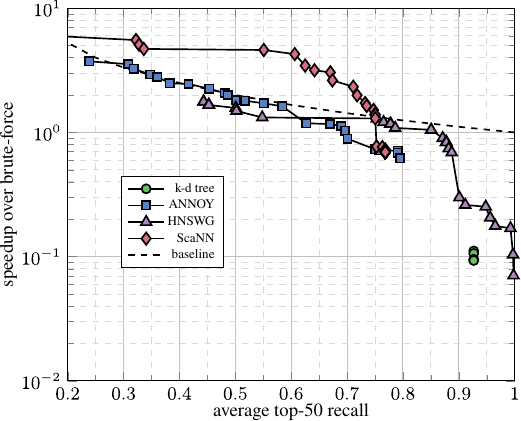}
    \end{minipage}
  \end{minipage} 
  \caption
    {
    Speedup (logarithmic scale) over brute-force search as a function of average top-50 recall, shown on logarithmic scale.
    The speedup is the ratio of time taken by brute-force search to the time taken by a given ANN method.
    Results obtained on the \textbf{Online Feature Learning (OFL)} dataset,
    starting with 5K samples, followed by 100K update events, event and search batch size of 200.
    \textbf{(a)}~Parameters for each method were tuned following recommended practice for static search problems.
    \textbf{(b)}~Parameters for each method were tuned via expanded parameter exploration in terms of range and resolution.
    }
  \label{fig:ofl_baseline_recall_runtime}
\end{figure}

The results are shown in Figs.~\ref{fig:odc_baseline_recall_runtime} and~\ref{fig:ofl_baseline_recall_runtime}
for the ODC and OFL datasets, respectively,
in terms of speedup over brute-force search across a range of accuracy points.
Accuracy is measured in terms of average recall,
where each search returns the \mbox{top-50} nearest neighbours
to a query and is scored against the ground truth \mbox{top-50}.
All neighbours are weighted equally for scoring.
This choice of neighbourhood size and weighting aims to approximate real-world application scenarios.
In addition to search runtime, the computation time for each method also includes \emph{event processing time}
(defined in Section~\ref{sec:characteristics}).

On both datasets,
the results first show that the best practice approaches for parameter tuning on static search problems 
do not directly translate to dynamic search problems,
resulting in the performance of many methods being well below the baseline.
Expanding the parameter exploration in terms of range and resolution,
in order to better accommodate the needs of dynamic search problems,
resulted in notably improved performance for all methods.

On the ODC dataset (Fig.~\ref{fig:odc_baseline_recall_runtime}),
using parameters tuned via expanded exploration, 
all methods except \mbox{k-d~trees} are faster than baseline,
while being capable of providing useful recall/runtime trade-offs.
HNSWG is the only method faster than the baseline above $70\%$ recall,
while the other methods do not reach $70\%$ recall.
Lastly, ScaNN is observed to be the fastest method below $70\%$ recall.

On the OFL dataset (Fig.~\ref{fig:ofl_baseline_recall_runtime}),
using parameters tuned via expanded exploration, 
all methods except ScaNN are either slower or on par with the baseline.
We conjecture that this is in part due to the \emph{update events} affecting all samples in an index,
while the \emph{addition events} (used in the ODC dataset) have a more local impact.
For recall below $75\%$, ScaNN outperforms the baseline.
Consistent with the ODC results, the k-d trees method is considerably slower than the baseline method.
We were unable to produce results for the current IVFPQ implementation due to memory constraints.

\clearpage
\subsection{Effect of Update Event Rates and Batch Sizes}

Update events can be quite disruptive in that they can affect all samples in an index.
To gauge the importance of update event rates as well as batch sizes
(defined in Section~\ref{sec:characteristics})
we perform experiments where are these aspects are varied.
For this task we use the HNSWG method on the OFL dataset,
as the results from Section~\ref{sec:comparative_eval} show that
the method is capable of operating over a wide range of recall points.
We evaluate two scenarios:
{(a)}~increasing update event rate with fixed search rate, 
and
{(b)}~increasing update event batch size with fixed search batch size.

\begin{figure}[!tb]
  \begin{minipage}{1\columnwidth}
    \begin{minipage}{0.05\textwidth}
      \centering
      \textbf{(a)} 
    \end{minipage}
    \hfill
    \begin{minipage}{0.925\textwidth}
      \centering
      \includegraphics[width=1\textwidth]{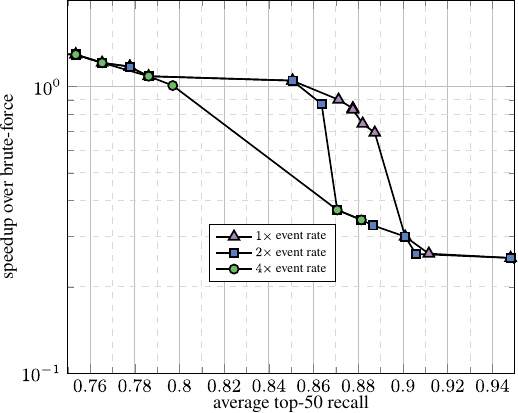}
    \end{minipage}
  \end{minipage}
  ~\\
  ~\\
  \begin{minipage}{1\columnwidth}
    \begin{minipage}{0.05\textwidth}
      \centering
      \textbf{(b)} 
    \end{minipage}
    \hfill
    \begin{minipage}{0.925\textwidth}
      \centering
      \includegraphics[width=1\textwidth]{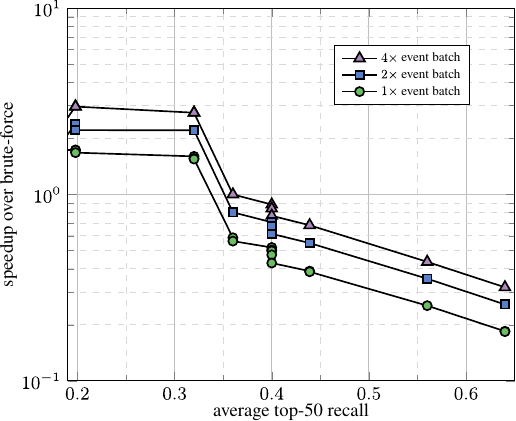}
    \end{minipage}
  \end{minipage} 
  \caption
    {
    Speedup over brute-force search as a function of average top-50 recall,
    while varying the rate and batch sizes of update events.
    Using the HNSWG method~\cite{malkov2018hnsw} on the OFL dataset,
    starting with 5k samples, followed by 100k update events, using initial event and search batch size of 200.
    For each subfigure, the axis limits were selected to zoom in on salient areas.
    \textbf{(a)}~Increasing event rate with fixed search rate.
    \textbf{(b)}~Increasing event batch size with fixed search batch size. 
    }
  \label{fig:frequency_batch}
  \vspace{0.5ex}
  \hrule
\end{figure}

The results shown in Fig.~\ref{fig:frequency_batch}
indicate that increasing the update event rate increases the difficulty of search on dynamic datasets,
with higher event rates tending to result in somewhat slower search times.
This is consistent with the view that more frequent update events within an allotted time period 
can reduce the quality of the index, thereby leading to slower search.
The results also show that increasing the update event batch size results in faster search times,
which is consistent with the stance that using more data for index updates can increase the quality of the index.

\vspace{1ex}
\section{Concluding Remarks}
\label{sec:conclusion}

Search on dynamic datasets is becoming more wide spread
due to increasing demands for fast and realtime processing of large data streams.
ANN methods can achieve sub-linear search times
by trading off between accuracy (recall) and runtime
through the use of various index structures to aid and accelerate the search process.

In contrast to search on static datasets,
the characteristics of search on dynamic datasets require taking into account
the computational costs of updating the index structure (to maintain search speed and accuracy),
as well as the rate and size of index updates.
Traditional approaches for evaluating ANN methods are focused on static datasets
and hence do not consider such characteristics,
which in turn leads to a lack of consensus as to which ANN methods are most suitable
for dynamic datasets.

In this work we have empirically evaluated 5 popular ANN methods 
(ScaNN~\cite{guo2020scann}, IVFPQ~\cite{johnson2019bf-ivfpq}, k-d trees~\cite{bentley1975kd}, ANNOY~\cite{bernhardsson2013annoy}, HNSWG~\cite{malkov2018hnsw})
while taking into account considerations pertinent to dynamic datasets.
Two dynamic datasets were used as proxies for broad real-world problems:
online data collection and online feature learning.
We used adapted forms of the 1~million sample SIFT1M dataset~\cite{sift1m,Jegou_2011} 
and the 1~billion sample DEEP1B dataset~\cite{babenko2016deep} 
as the dynamic datasets.

The results first show that best practice approaches for parameter tuning based on static search problems
do not directly translate to dynamic search problems,
as the performance of many methods was observed to be below a straightforward baseline exhaustive search method.
While expanding the parameter exploration in terms of range and resolution
(to better accommodate the needs of dynamic search problems)
resulted in notably improved performance,
the improvements were not necessarily sufficient.
To wit, the k-d trees method appears not to be suitable for dynamic datasets
as it obtained performance worse than baseline even after the expanded parameter exploration,
despite being previously applied on such datasets~\cite{kim2018useskd,xu2021useskd}.

The results further show that for online data collection
the HNSWG method achieves a consistent speedup over the baseline method across a wide range of recall rates.
For online feature learning, the ScaNN method is faster than the baseline method for recall rates below 75\%.
The results also indicate that batching a larger number of samples into index updates
can increase the relative speed of the HNSWG method for online feature learning.
This suggests that other ANN methods with slow index construction may also benefit
from reduced update frequencies and increased batch sizes.

\pagebreak
Future avenues of research include studying the effects of pruning a dataset
(eg.~removal of old or low saliency samples),
which is an important consideration for bounding memory use.
Furthermore, it would be useful to evaluate the effects of temporal dependence (mutual information)
between consecutive samples due to temporal consistency in applications such as robotic localisation~\cite{garg2021seqnet}.
Exploiting mutual information between consecutive search batches
may provide large performance gains, 
due to the possibility of basing the starting point of each search
on the search results obtained for preceding samples.
Lastly, it may be beneficial to explore the trade-offs
between accuracy and explainability of each ANN method,
in order to take into account aspects of responsible~AI~\cite{Sanderson_2023,Sanderson_2024}.

\vspace{6ex}
\balance
\bibliographystyle{ieee_mod}
\bibliography{references}

\end{document}